\documentclass{article}

\usepackage{microtype}
\usepackage{graphicx}
\usepackage{booktabs}
\usepackage{hyperref}

\usepackage[accepted]{icml2026}

\usepackage{amsmath}
\usepackage{amssymb}
\usepackage{algorithm}
\usepackage{algorithmic}
\usepackage{enumitem}
\usepackage{xcolor}
\usepackage[capitalize,noabbrev]{cleveref}
\usepackage{tabularx}

\icmltitlerunning{Adaptive Minds: Empowering Agents with LoRAs as Tools}

\begin{document}

\twocolumn[
  \icmltitle{Adaptive Minds: Empowering Agents with LoRAs as Tools}

  \begin{icmlauthorlist}
    \icmlauthor{Pavan C Shekar}{qpiai}
    \icmlauthor{Aswanth Krishnan}{qpiai}
  \end{icmlauthorlist}

  \icmlaffiliation{qpiai}{QpiAI, Bengaluru, India}
  \icmlcorrespondingauthor{Pavan C Shekar}{pavan.s@qpiai.tech}
  \icmlcorrespondingauthor{Aswanth Krishnan}{Ashwanth.krishnan@qpiai.tech}

  \icmlkeywords{compositional generalization, LoRA-as-Tools, Multi-step agentic reasoning, ReAct agents,
    multi-adapter orchestration, interpretability, parameter-efficient fine-tuning}

  \vskip 0.3in
]

\printAffiliationsAndNotice{\textit{Accepted to the CompLearn Workshop at ICML 2026 (non-archival).}}

\begin{abstract}
We investigate a framework in which LoRA adapters are treated as callable tools that a base language model can dynamically select and invoke. We hypothesize that, when adapters are trained to provide strong domain-specific gains and are exposed with clear metadata, a base model can reliably route queries to the appropriate expert, effectively aggregating the benefits of many specialized adapters within a single framework. We introduce Adaptive Minds, a general framework within which we study both single-step routing and multi-step agentic reasoning. In this setting, the agent can iteratively invoke multiple adapters alongside other tools (e.g., external APIs, retrieval systems, or execution environments) and reason over their outputs across multiple steps. This reframes adapters as modular skills or memory units that can be composed during reasoning rather than statically applied. In our evaluation, the routing layer reaches 98.3\% accuracy on a 30-adapter library, and well-trained specialists provide $+4.6$ to $+84.0$ percentage points of strict-scorer gain across nine task families under a single shared training recipe; the AM router aggregates these gains within $\pm 5$\,pp of the direct specialist on every benchmark whose queries surface domain signal. Our findings suggest that the effectiveness of this approach depends on the quality and specialization of individual adapters, and that enabling flexible composition of many such experts can significantly expand the practical capabilities of language model agents, moving toward more general, tool-augmented intelligence.\footnote{Code: \href{https://github.com/qpiai/adaptive-minds}{\texttt{github.com/qpiai/adaptive-minds}}}
\end{abstract}

\section{Introduction}
Recent advances in language models have produced two powerful but largely disconnected capabilities: parameter-efficient specialization through adapters~\citep{houlsby2019parameter,hu2022lora}, and tool-augmented reasoning through agents~\citep{yao2023react,karpas2022mrkl}. LoRA adapters are lightweight parameter updates that specialize a shared base model without requiring a full copy of the model -- a property driving the rise of domain-specialist LLMs in medicine, finance, and law~\citep{singhal2023clinical,cheng2024adaptllm}, while agentic systems improve performance by interleaving reasoning with actions such as retrieval, execution, and API use. In existing multi-adapter workflows, however, selecting which adapter to use is still often handled manually or through fixed heuristics. This raises a natural question: can adapter selection itself become part of the model's reasoning process?

We introduce Adaptive Minds, a unified framework that answers this question by treating LoRA adapters as tools that a base model can dynamically select and compose. At one end, this yields semantic routing, where the model chooses the appropriate specialized adapter from metadata and query semantics, replacing manual dispatch and brittle keyword matching with model-driven selection. At the other, the same formulation extends to multi-step agentic reasoning, where adapters can be invoked alongside other tools to solve more complex tasks requiring decomposition, intermediate reasoning, or multiple forms of expertise. This reframes adapters from static fine-tuning artifacts into reusable capability units that can be called and combined during inference. Like any other tool an agent reaches for, a LoRA adapter is bounded by how well it is built. Across our experiments we observe all three outcomes from the same framework: well-trained specialists deliver large per-domain gains that the router automatically aggregates, weakly differentiated adapters leave accuracy unchanged, and in some cases adapters can degrade accuracy below the base model. The user-facing implication is simple --- train good adapters, and the framework takes care of when, where, and how to use them. More broadly, the framework points toward a different route to stronger and more general model behavior: not only by scaling a single monolithic model, but by enabling a base model to orchestrate many specialized capabilities within a single interpretable framework. This explicit, modular composition complements a long line of work showing that monolithic neural models often struggle with systematic compositional generalization~\citep{hupkes2020compositionality,lake2018generalization}.

\section{Contributions}
\textbf{(1)} We present \emph{Adaptive Minds}, a general framework in which LoRA adapters are exposed as tools that a base model can dynamically select and compose.

\textbf{(2)} We unify two levels of adaptation within this framework: semantic routing over a library of domain specialists, and agentic multi-step reasoning in which multiple adapters can be used alongside external tools to solve more complex tasks.

\textbf{(3)} We demonstrate that the same framework supports both direct adapter routing and multi-step agentic reasoning, enabling models to invoke specialized adapters together with external tools across multiple reasoning steps.

\textbf{(4)} We measure that, under one shared training recipe and a single backbone, well-targeted LoRA specialists improve strict-scorer accuracy by $+4.6$ to $+84.0$\,pp across nine task families, and the AM router aggregates these gains within $\pm 5$\,pp of direct expert selection on every benchmark whose queries carry domain-discriminating signal (Section~\ref{sec:conditionality}).

\section{Related Work}
\label{sec:related}

\subsection{Tool-Augmented Agents}

Tool-augmented language models improve performance by interleaving reasoning with actions such as retrieval, execution, or API use. ReAct~\citep{yao2023react} established the reasoning--action loop, while Toolformer~\citep{schick2024toolformer} and Gorilla~\citep{patil2024gorilla} broadened the view of tools as modular interfaces that language models can invoke. HuggingGPT~\citep{shen2023hugginggpt} extended this view further by treating entire pre-trained models as callable tools that an LLM controller can plan over, Reflexion~\citep{shinn2023reflexion} introduced verbal self-reflection, and Tree-of-Thoughts~\citep{yao2023tot} extended reasoning to deliberate multi-path exploration, both refining agent behavior over reasoning steps. In these systems, however, tools are typically external capabilities. Our setting differs in that we ask whether a model's own fine-tuned specializations can also be exposed as tools, allowing LoRA adapters to become resources that an agent can explicitly select and use during inference.

\subsection{LoRA Routing and Serving}
A growing body of work studies how to host, switch, and route among many LoRA adapters efficiently. Systems such as S-LoRA~\citep{sheng2024slora} and dLoRA~\citep{wu2024dlora} focus primarily on efficient multi-adapter serving, while LoRA-Switch and related approaches perform finer-grained routing or blending, often at the token level. Recent methods such as LoRAMoE~\citep{dou2024loramoe}, MoLoRA~\citep{molora2026}, and task-level adapter routing approaches further show that inference-time routing can scale to large adapter libraries. This line of work is closely related to ours, but its primary emphasis is on efficient serving, blending, or adapter selection. Our framework operates at a different layer and is composable with these systems --- we focus on treating adapters as explicit tools inside a reasoning process, where the model can decide not only \emph{which} adapter to use, but also \emph{when}, \emph{how often}, and \emph{in what sequence} to invoke multiple adapters while solving a problem. Our own implementation, for example, uses vLLM~\citep{kwon2023vllm} as the underlying serving stack.

A closely related line of work learns a continuous gating router over LoRA experts at training time, often inside the transformer itself~\citep{dou2024loramoe}. Our framework is conceptually orthogonal: selection is performed by the base model at the tool-call level using natural-language metadata, requiring no additional router training and exposing each adapter invocation as a discrete, named action. This trades the optimization capacity of learned gating for interpretability, modularity, and the ability to add or remove adapters at deployment time without re-training a router.

\subsection{LoRA Composition}
Another related direction studies composition at the parameter level, drawing on the task-arithmetic paradigm in which fine-tuning deltas are treated as composable vectors~\citep{ilharco2023editing,yadav2023tiesmerging}. LoRAHub~\citep{huang2024lorahub} formulates dynamic composition through adapter selection and weighting, while LoRA Soups~\citep{prabhakar2025lorasoups}, building on weight-averaging foundations~\citep{wortsman2022modelsoups}, explores merging LoRAs into a single combined parameter state. Our framework is complementary but conceptually different: the primary form of composition happens across reasoning steps and intermediate outputs rather than only within merged parameters. In our setting, composition is an agentic process rather than purely a weight-space operation, making adapter usage more explicit, interpretable, and easier to analyze.

\subsection{Modular Computation and Expert Routing}

Our work is also connected to research on modular computation and expert routing. Neural Module Networks~\citep{andreas2016nmn} show that complex behavior can emerge by routing inputs through specialized sub-networks, and Routing Networks~\citep{rosenbaum2018routing} extend this paradigm to multi-task learning by jointly training a router with a set of function blocks. Mixture-of-Experts~\citep{jacobs1991adaptive,shazeer2017moe} extends this intuition to learned expert routing, typically at the token or layer level. We draw on this literature but shift the unit of specialization to the adapter level. Unlike MoE architectures, which rely on a fixed set of learned experts inside a single model, our framework treats adapters as modular inference-time tools that can be added, removed, and composed more flexibly. This also makes the framework compatible with a wide range of base models, including architectures that may themselves already use MoE internally.

Across these lines of work, prior research has shown how to build agents with external tools, how to serve and route many LoRA adapters efficiently, and how to merge adapters at the parameter level. Our contribution sits at the intersection of these directions. We treat LoRA adapters as tools inside a general reasoning framework. This allows a base model to perform both semantic routing for simple requests and multi-step consultation for more complex ones, while preserving a clearer execution trace than purely implicit parameter-level composition.

\section{LoRA-as-Tools: Formalism}
\label{sec:formalism}

We formalize each LoRA adapter as a tool that can be selected and invoked by a
base model during inference.
Given an adapter $a_i$ with parameters $\Delta\theta_i$, a domain description
$d_i$, and an associated prompt template $s_i$, we define the corresponding
tool as
\begin{equation}
\tau_i = (\texttt{name}_i,\ d_i,\ \{\texttt{sub\_query} : \text{str}\},\ \texttt{generate}_i),
\end{equation}
where $\texttt{generate}_i$ activates adapter $a_i$ via
\texttt{set\_adapter}$(a_i)$, constructs an expert-conditioned prompt from
$s_i$, and returns the resulting domain-specific response.

This formulation turns adapters from static fine-tuning artifacts into explicit
computational resources that can be planned over during reasoning.
The tool registry is generated directly from adapter metadata, so adding a new
domain expert requires no changes to the agent logic.

Under this view, routing appears as the special case of tool use in which only
a single expert is consulted.
Let $Q$ denote the user query, $\mathcal{H}_{<k}$ the interaction history up to
step $k$, and $M_\theta$ the base model.
At each step, the model reasons over the query and prior context,
selects an expert if needed, and obtains an observation:
\begin{align}
t_k &= \text{Think}(Q, \mathcal{H}_{<k}; M_\theta), \\
o_k &= \text{Generate}(\text{sq}_k; M_{\theta + \Delta\theta_{j_k}}), \\
R &= \text{Synthesize}(Q, \{o_1, \ldots, o_K\}; M_\theta).
\end{align}
When $K=1$, this reduces to semantic routing.
When $K>1$, the model can iteratively consult multiple experts and combine their
outputs across reasoning steps.

An important consequence of this formulation is auditability.
Because each tool invocation is an explicit, named action, the resulting
composition trace $\{(t_k, j_k, \text{sq}_k, o_k)\}_{k=1}^{K}$ is observable
and attributable.
This provides a clearer account of how expertise is used than parameter-level
composition, where the contribution of individual adapters is entangled in the
merged weights.

\section{System Architecture}
\label{sec:architecture}

\subsection{Adaptive Minds Framework}

Adaptive Minds is a unified framework that supports both expert routing and multi-step agentic reasoning.
The central idea is simple. Adapters are exposed as tools, and the base model
decides whether a query can be handled by consulting a single expert or whether
it requires a richer sequence of reasoning steps involving multiple experts and
other tools.
Rather than treating routing and agentic reasoning as separate paradigms, we
view them as two operating regimes of the same architecture.

\begin{figure}[t]
\centering
\includegraphics[width=\columnwidth]{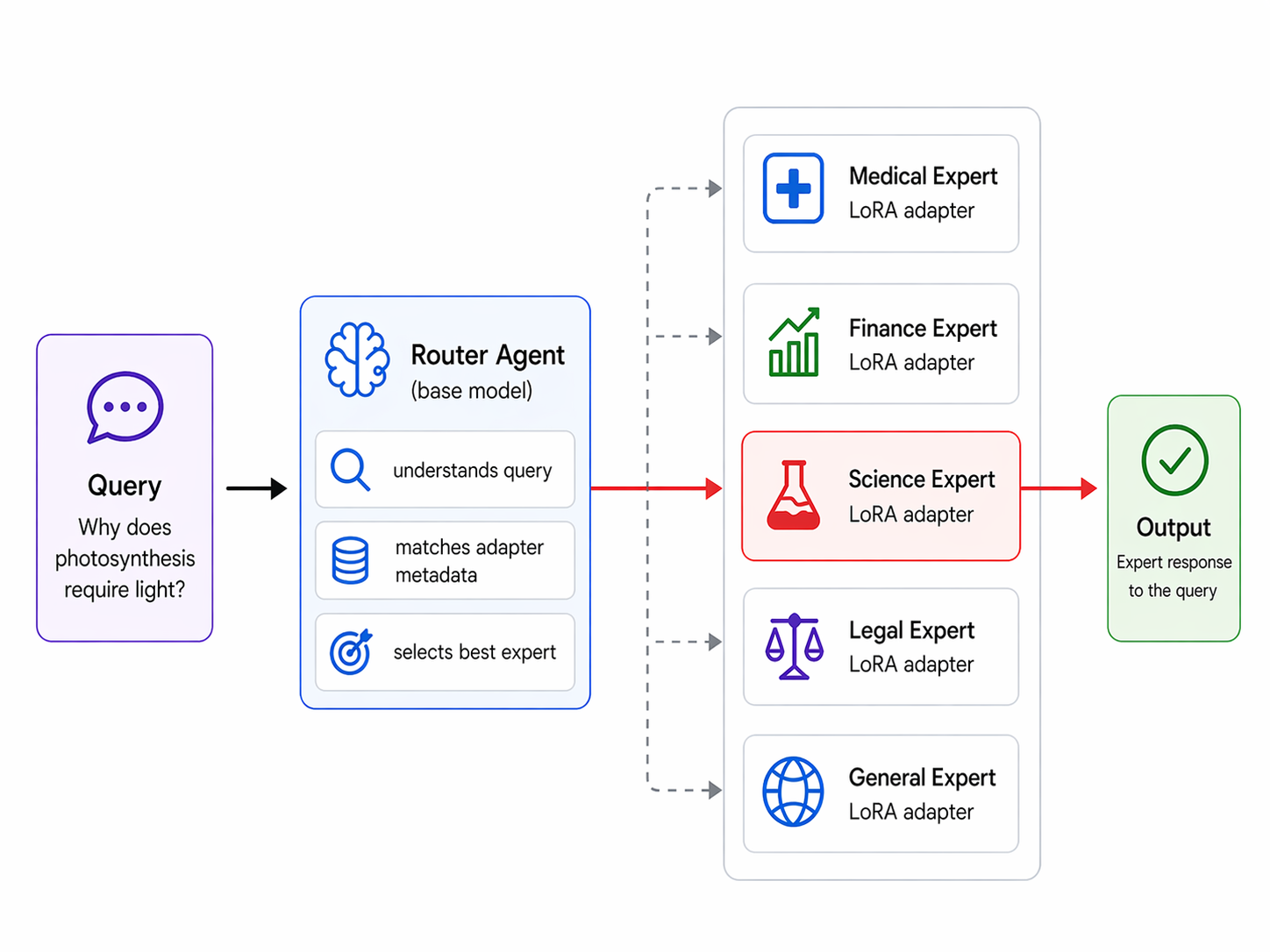}
\caption{Routing architecture of Adaptive Minds. Given a user query, the base model acts as a router that matches the query to the best domain-specialized LoRA adapter and dispatches it for expert execution. This provides a simple single-step operating mode that replaces manual adapter selection with interpretable model-driven routing.}
\label{fig:arch}
\end{figure}

\subsection{From Routing to Agentic Reasoning}

At one end of this framework is semantic routing.
For queries that are primarily single-domain, the model selects the most
appropriate adapter from its metadata and dispatches the query directly to that
expert.
This provides a practical way to aggregate the benefits of many specialized
LoRAs within a single deployed framework, without relying on brittle keyword rules
or hand-written dispatch logic.

For more complex queries, the framework naturally extends to an agentic
setting.
Here the model can iteratively invoke multiple adapters alongside external
tools such as retrieval, search, execution, or other task-specific resources.
This allows the framework to decompose a problem into sub-queries, consult
different forms of expertise when needed, and synthesize a final answer from
intermediate observations.
In this regime, adapters function not only as specialists, but also as modular
skills or memory-like units that can be called and combined during reasoning.

\begin{figure}[t]
\centering
\includegraphics[width=\columnwidth]{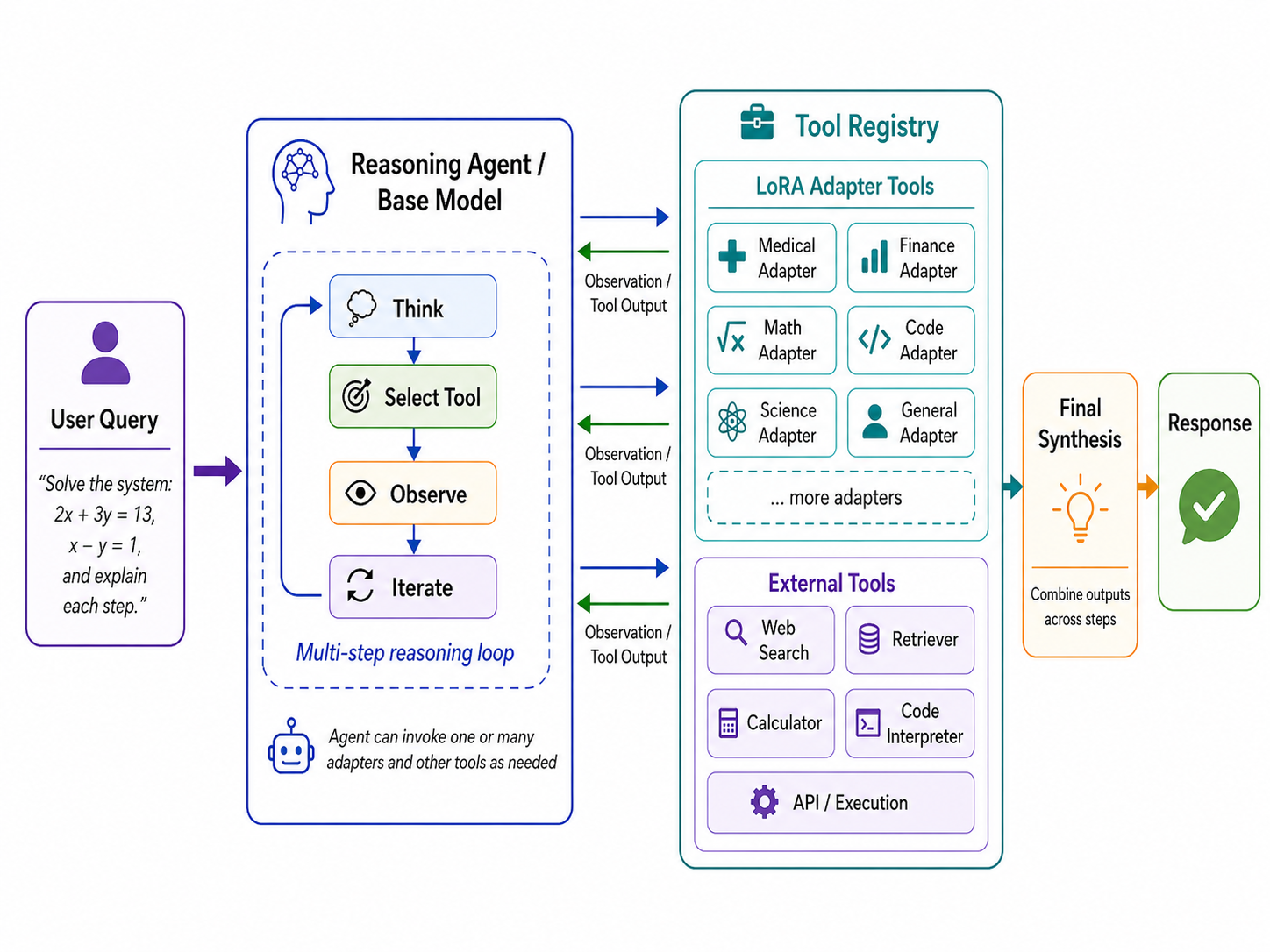}
\caption{Multi-step agentic reasoning in our framework. LoRA adapters are exposed as callable tools alongside standard external tools, allowing the base model to perform iterative tool selection, observation, and final synthesis over multiple reasoning steps.}
\label{fig:agent-arch}
\end{figure}

\begin{figure}[t]
\centering
\includegraphics[width=\columnwidth]{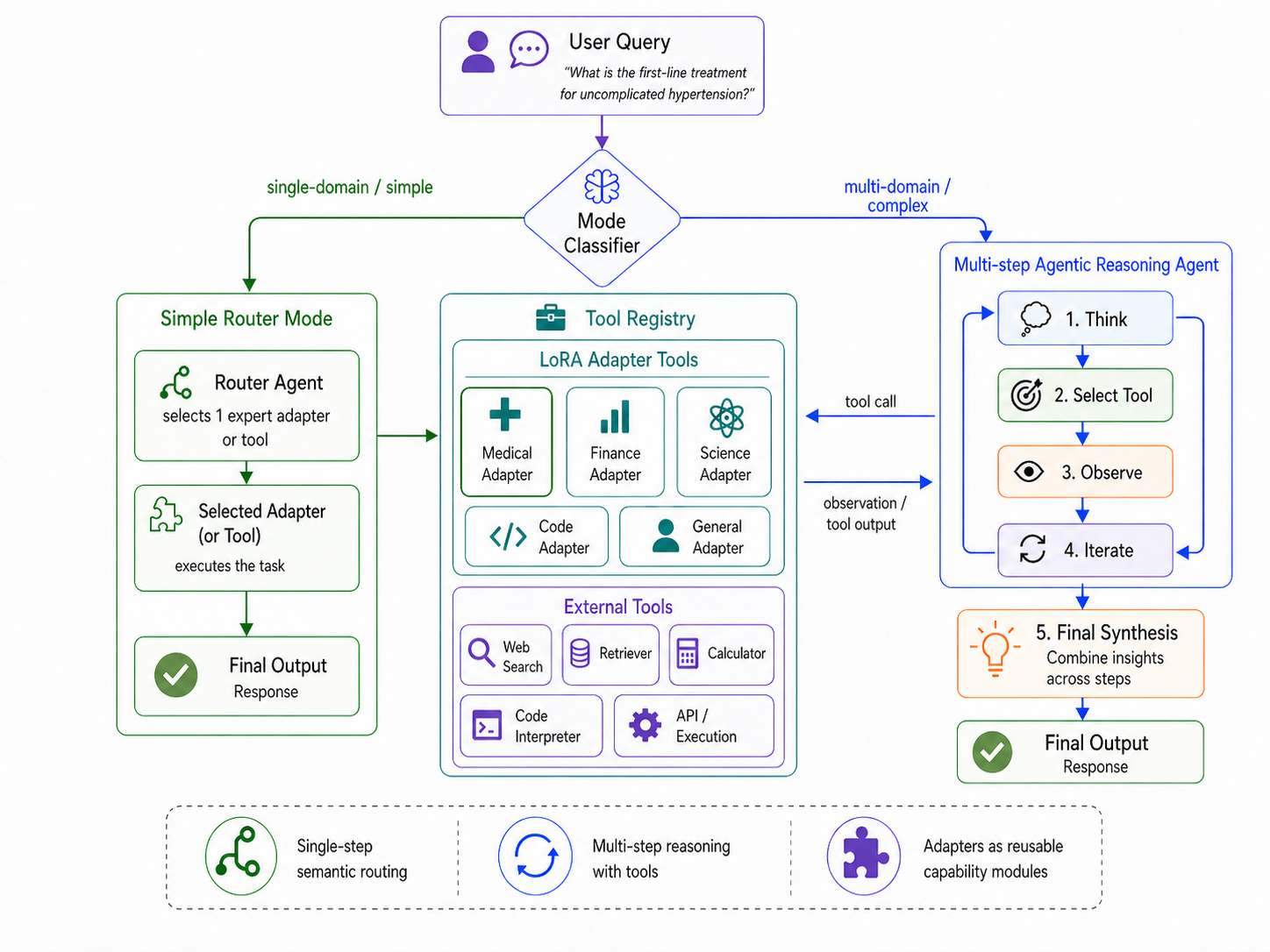}
\caption{Unified operating modes of the framework. The system first classifies query complexity, then either dispatches the query through single-step semantic routing or invokes a multi-step agentic reasoning loop that composes specialized adapters with external tools.}
\label{fig:dualmode}
\end{figure}

\subsection{Unified Adapter Runtime}

We implemented the framework over two complementary runtimes for multi-adapter inference. The first is a direct PEFT-based runtime in which all specialists are attached to a shared base model and selected dynamically at inference time. This setting allowed us to study adapter switching and placement strategies directly, including GPU-resident, CPU-staged, and on-demand loading. The second runtime, serves the full adapter library through a single vLLM engine with LoRA support~\citep{kwon2023vllm}. In this setting, expert selection is handled at request time through the serving interface, while the runtime manages adapter paging and execution internally. Together, these implementations show that our approach is not tied to a single serving stack, but can be realized across different multi-adapter runtimes.

A key architectural principle is that control and reasoning remain with the base model, while specialized adapters are used only for expert execution. This separation preserves the general reasoning ability of the base model while allowing domain-specific capabilities to be injected only when needed. As a result, the framework maintains a clear distinction between deciding \emph{what} expertise is required and generating \emph{with} that expertise, making adapter use more modular, more interpretable, and easier to audit than implicit parameter-level composition. In practice, we found the vLLM-based runtime to be both faster and operationally cleaner for large-scale multi-adapter serving. It hosted all 30 adapters within a single engine and achieved substantially lower router-mode latency than the earlier PEFT-based runtime, while preserving the same high-level routing abstraction.

\subsection{Agent Loop}

The agent operates by alternating between reasoning and tool use.
Given a query $Q$ and a tool registry $\mathcal{T}$, the base model first
reasons about the problem and decides whether an action is needed.
If the model selects an adapter tool, the corresponding specialist is activated,
a sub-query is executed, and the resulting observation is appended to the
interaction history.
This process can repeat for multiple steps before the base model synthesizes a
final answer from the accumulated observations.

This loop captures the general case of adapter composition.
When only one expert is selected, the process collapses to routing; when
multiple steps are required, the same mechanism supports iterative
multi-step reasoning.

\subsection{Dynamic Control and Verification}
\label{sec:control}

To support this spectrum of behaviors, the framework includes lightweight
control mechanisms that determine when simple routing is sufficient and when
full multi-step reasoning is required.
In our implementation, we score query difficulty by the mean predictive
entropy of the base model over the first $T{=}16$ tokens it would emit in
response to the query. Let $p_t(v) \triangleq p_t(v \mid Q, h_{<t})$; then
\begin{equation}
\label{eq:entropy-gate}
H(Q) \;=\; \frac{1}{T}\sum_{t=1}^{T}\Bigl(-\!\!\sum_{v\in\mathcal{V}} p_t(v)\,\log p_t(v)\Bigr),
\end{equation}
where $\mathcal{V}$ is the model's vocabulary, $p_t(v)$ is the next-token
distribution at step $t$ conditioned on the query and the model's own decoded
prefix $h_{<t}$.
Queries with $H(Q) < 0.8$ are routed to single-adapter mode (Router);
queries with $H(Q) > 1.5$ are routed to multi-step mode (Agent); and
intermediate scores fall back to Router with the option for the base model
to escalate during reasoning. These thresholds are configurable and were
selected once on internal validation queries and held fixed across all
reported experiments; a full sensitivity analysis is left to future
work. We additionally include a verification stage that evaluates
intermediate observations before they are incorporated into the final
synthesis (Appendix~\ref{app:agent-details}), reducing the risk that weak
expert outputs propagate through the reasoning chain.

More broadly, these mechanisms are not tied to a particular prompting strategy
or tool set.
They reflect a general design principle. Adapter composition should remain
selective, interpretable, and grounded in the evolving needs of the query,
rather than being applied uniformly to every input.

\section{Domain Experts}
\label{sec:experts}

We train domain-specialized LoRA adapters on Qwen2.5-7B-Instruct
and Qwen3.5-9B~\citep{qwen3techreport} under one shared recipe per
track (reasoning-trace SFT, structured-generation SFT, or GRPO
with execution-based reward), so that downstream behavioral
differences reflect domain specialization rather than
hyperparameter variation. Each adapter is lightweight relative
to the base model, making it practical to host many experts
simultaneously and switch among them at inference time. Full
per-track recipes are reported in Appendix~\ref{app:inventory}.

\begin{table}[t]
\caption{Routing accuracy as the adapter pool scales from 5 to 30 adapters. The 5-adapter evaluation uses Llama~3.1~8B on 29 queries; the 30-adapter evaluation uses Qwen3.5-9B on 60 queries.}
\label{tab:routing}
\centering
\small
\setlength{\tabcolsep}{3pt}
\begin{tabular}{@{}lccccc@{}}
\toprule
\textbf{Model} & \textbf{Adapters} & \textbf{\# Queries} & \textbf{Keyword} & \textbf{Router} & \textbf{$\Delta$} \\
\midrule
Llama~3.1~8B & 5  & 29 & 48.3 & \textbf{100.0} & +51.7 \\
Qwen3.5-9B   & 30 & 60 & 31.7 & \textbf{98.3}  & +66.6 \\
\bottomrule
\end{tabular}
\end{table}

\begin{table*}[t]
\caption{Consolidated specialist results on Qwen2.5-7B-Instruct (one shared training recipe, all adapters served by a single vLLM engine). Strict-scorer accuracy per benchmark; AM denotes the deployed router (heuristic + LLM fallback); Router-acc is the fraction routed to the domain-matched specialist; ``--'' = AM not run. $^\dagger$AM on \texttt{nl2bash} measured on $n{=}100$ subset (specialist + base direct on full $n{=}606$). $^\ddagger$On benchmarks where the base scores 0\% (LEDGAR), the strict scorer rejects structurally adjacent but non-conformant outputs from the base (e.g., emitting ``E1'' or ``R-WRepWarranty'' on LEDGAR rather than the integer label ``98''); the specialist learns the target output convention. The headline gain on such rows is therefore best read as \emph{format acquisition} rather than knowledge acquisition; see Appendix~\ref{app:ledgar-format} for worked examples.}
\label{tab:specialists}
\centering
\small
\setlength{\tabcolsep}{6pt}
\begin{tabular}{@{}lrrlrrr@{}}
\toprule
\textbf{Benchmark} & $n$ & \textbf{Base} & \textbf{Specialist} & \textbf{AM} & \textbf{Router-acc} & \textbf{$\Delta$ Specialist} \\
\midrule
\multicolumn{7}{l}{\emph{Strong-base reasoning (no capability gap)}} \\
MATH-500 & 500 & 69.2\% & 67.2\% (reasoning v1) & -- & -- & $-2.0$ pp \\
GSM8K-250 & 250 & 86.0\% & 88.0\% (reasoning v1) & -- & -- & $+2.0$ pp \\
\midrule
\multicolumn{7}{l}{\emph{Weak-base structured generation (five grammars under one recipe)}} \\
Spider-dev (SQL) & 300 & 12.3\% & 41.7\% (sql v1) & 39.7\% & 100\% & $+29.4$ pp \\
Text2Cypher (graph queries) & 300 & 0.0\% & 39.3\% (cypher v1) & 38.0\% & 92\% & $+39.3$ pp \\
nl2bash (shell pipelines) & 606 & 8.4\% & 17.3\% (bash v1) & 12.0\%$^\dagger$ & 98\% & $+8.9$ pp \\
LC-QuAD (SPARQL, structural) & 100 & 0.0\% & 59.0\% (sparql v1) & 58.0\% & 95\% & $+59.0$ pp \\
MermaidSeqBench (visual DSL) & 100 & 4.5\% & 21.2\% (mermaid v1) & 24.0\% & 100\% & $+16.7$ pp \\
\midrule
\multicolumn{7}{l}{\emph{Tool-style and niche adapters}} \\
PII (NER+redaction) & 100 & 2.7\% & 79.0\% (pii v1) & 74.0\% & 99\% & $+76.3$ pp \\
Qiskit-HumanEval-hard (quantum, GRPO) & 151 & 10.6\% & 15.2\% (quantum v1) & 15.0\% & 93\% & $+4.6$ pp \\
LEDGAR (legal, 3-stage curriculum)$^\ddagger$ & 100 & 0.0\% & 84.0\% (legal v1) & 84.0\% & 100\% & $+84.0$ pp \\
ChEMBL canonical SMILES (chem) & 300 & 6.3\% & 36.7\% (chem v1) & 35.0\% & 100\% & $+30.4$ pp \\
\bottomrule
\end{tabular}
\end{table*}

\section{Experiments}
\label{sec:experiments}

\subsection{Experimental Setup}

We evaluate our framework along two complementary axes.
First, we study \emph{semantic routing}: whether a base model can
reliably select the appropriate domain-specialized adapter from
metadata and query semantics. Second, we study \emph{specialist-driven
gains and their aggregation}: whether well-targeted LoRA adapters
trained under one shared recipe provide measurable gains on weak-base
structured-generation and niche tasks, and whether the AM router
recovers those gains automatically within a single deployed framework.

Experiments run on a single NVIDIA L40S (48\,GB). The consolidated
specialist study uses Qwen2.5-7B-Instruct; routing experiments use
Qwen3.5-9B and Llama~3.1~8B as noted. A complementary MCQ evaluation
of four reasoning-trace adapters is in Appendix~\ref{app:domain-eval}.
All Qwen evaluations use \texttt{enable\_thinking=False}, greedy
decoding, and Wilson 95\% confidence intervals.

\subsection{Semantic Routing Accuracy}

Table~\ref{tab:routing} shows that model-driven routing remains highly reliable as the adapter library grows substantially in both size and ambiguity. In the 5-adapter setting, the router achieves \textbf{100.0\%} accuracy, outperforming keyword matching by \textbf{51.7} percentage points. In the 30-adapter setting, it still achieves \textbf{98.3\%} accuracy on a 60-query manually curated evaluation set, despite a much larger choice space and genuine overlap between related adapter families. For example, prompts in mathematics may reasonably map to \textbf{math}, \textbf{math\_reasoning}, or \textbf{math\_grpo}, while prompts in medicine may overlap across \textbf{medical}, \textbf{biomedical}, and \textbf{medical\_reasoning}. In this larger setting, some queries admit more than one valid adapter because several specializations cover closely related capabilities; predictions are therefore counted as correct when the selected adapter falls within the valid set for that query. By contrast, the keyword baseline drops from \textbf{48.3\%} to \textbf{31.7\%}, increasing the router's absolute advantage from \textbf{+51.7} to \textbf{+66.6} percentage points.

This scaling result is important because it tests routing under a more realistic condition than the earlier small-adapter setting. With overlapping adapters, the problem is no longer simply whether the router can match a query to a single isolated specialization, but whether it can make a sensible choice among multiple nearby capabilities. The results suggest that it can: accuracy remains effectively intact even as the adapter pool grows by $6\times$ and the routing boundaries become less keyword-like and more semantic. This supports our central claim that, when adapters are exposed with clear metadata, the base model itself can serve as an effective router over a large library of specialized adapters, making model-driven selection a viable alternative to manual switching or brittle keyword heuristics.

\paragraph{When routing works: a mechanism, not a uniform property.}
Per-benchmark routing accuracy varies sharply with how much
adapter-specific surface signal the query distribution retains. On
benchmarks whose queries naturally contain trigger
vocabulary --- PII (\textit{``redact''}), Bash (\textit{``grep''},
\textit{``sed''}), Mermaid (\textit{``sequence diagram''}), Regex
(\textit{``regex''}) --- the router places 93--97\% of queries on
the domain-matched specialist, closely tracking the curated
98.3\%. As a positive spot-check on naturalistic surface form, we
also evaluate routing on PII redaction over benign-looking personal
letters (e.g., \textit{``Dear Haylee, I see you're reaching your
90th birthday March 15, 1941\ldots''}) and find the deployed router
still routes correctly on 93\% of $n{=}300$ such queries. On
benchmarks whose queries do not surface specialist need ---
LC-QuAD SPARQL questions phrased as natural English
(\textit{``Who was Caspar David Friedrich's student?''}, $0/8$ of
the SPARQL adapter's trigger keywords present) or function-calling
intents phrased without explicit \textit{``call''} / \textit{``JSON''}
/ \textit{``API''} markers ($0/6$ of the function-call adapter's
keywords present) --- the router most often defers to the base
model. This is consistent with the framework's design that places
control and reasoning with the base model and reserves specialists
for queries that explicitly invoke their expertise: a user asking
\textit{``who was Friedrich's student''} wants a natural-language
answer, not a SPARQL program, so base-model deferral is the correct
deployment behavior on such queries. A residual failure mode is
queries that route to a shape-similar but wrong specialist (e.g.,
a SQL specialist on relational-shaped function-calling queries),
which bounds the deferral benefit. Two practical consequences
follow. First, per-benchmark routing rate at deployment is
predictable in advance from a simple keyword-overlap diagnostic
between adapter metadata and the target query distribution. Second,
for deployments that need the framework to actively detect
specialist applicability from query \emph{structure} rather than
surface vocabulary, richer adapter metadata or a learned router
fine-tuned on the deployment distribution are natural extensions.
We expand on the per-benchmark gradient in
Appendix~\ref{app:routing-real}.

\paragraph{Deployment cost.}
On the 30-adapter library, vLLM serves the full set from a single
engine at mean router-mode latency 3.48\,s on a single L40S; adapter
placement (GPU-resident / CPU-staged / on-demand) changes end-to-end
latency by $<\!5\%$, so generation time dominates
(Appendix~\ref{app:latency}).

\subsection{Specialists Drive the Gains; AM Aggregates Them}
\label{sec:conditionality}

A consolidated study of nine task-matched specialists on weak-base structured-generation and niche tasks reveals the empirical core of the framework. Table~\ref{tab:specialists} reports per-benchmark base accuracy, accuracy with the domain-matched specialist directly pinned, accuracy of the deployed AM router that selects the specialist via metadata, and the router-acc fraction giving the rate at which the router actually dispatched to the domain-matched adapter. A complementary MCQ evaluation of four reasoning-trace adapters on Qwen3.5-9B (Appendix~\ref{app:domain-eval}) shows the same conditional pattern in miniature: only Finance $\times$ FinanceBench produces a clean $+4.0$\,pp in-domain gain.

Three patterns in Table~\ref{tab:specialists} are the empirical core of the paper. \textbf{(i) Specialists help where the base has a real capability gap.} On strong-base reasoning tasks (MATH-500, GSM8K), the recipe moves accuracy by less than $\pm 2$\,pp, within seed-to-seed noise. On the nine weak-base structured-generation and niche tasks, specialists improve strict-scorer accuracy by $+4.6$ to $+84.0$\,pp under the same shared training recipe. \textbf{(ii) AM aggregates these gains.} AM-route tracks the directly-pinned specialist within $\pm 5$\,pp on all 9 measured benchmarks --- the router recovers each per-domain gain without a human picking the right adapter. \textbf{(iii) Routing is reliable at scale.} Combined with the routing study above (98.3\% on a 30-adapter pool, $+66.6$\,pp over keyword matching), the pipeline of trained adapters $+$ reliable router $+$ per-specialist accuracy is coherent end-to-end.

\paragraph{Directional transfer.}
We additionally measured how each specialist performs on every other domain's benchmark, to test whether the gains reflect domain-matched expertise rather than a generic ``structured-output'' lift. Off-domain transfer is essentially zero on 7 of 9 specialist-vs-benchmark pairs. Two pairs show partial transfer with a plausible structural explanation: a Bash specialist scores above the regex specialist on a regex benchmark (reflecting \texttt{grep}/\texttt{sed} patterns in \texttt{nl2bash} training data), and a SQL specialist matches the Mermaid specialist on Mermaid sequence diagrams (reflecting shared arrow-like structured-token patterns). The domain-matched-expertise claim therefore holds modulo structurally adjacent grammars.

\subsection{Routing vs.\ Multi-Step Reasoning}

To better understand the regimes in which routing and iterative reasoning are useful, we compare vanilla inference, semantic routing, and multi-step agentic reasoning on 100 domain-matched MMLU questions using Llama~3.1~8B.

\begin{table}[t]
\caption{Three-way comparison on 100 domain-matched MMLU questions (Llama~3.1~8B).}
\label{tab:3way-main}
\centering
\small
\begin{tabular}{@{}lcccccc@{}}
\toprule
\textbf{Mode} & \textbf{All} & \textbf{Chem} & \textbf{Fin} & \textbf{AI} & \textbf{Med} & \textbf{Gen} \\
\midrule
Vanilla & 56.0 & 55 & 65 & 25 & 80 & 55 \\
\textbf{Router} & \textbf{62.0} & 60 & 60 & 40 & 75 & \textbf{75} \\
Agent & 58.0 & \textbf{70} & 40 & \textbf{45} & 75 & 60 \\
\bottomrule
\end{tabular}
\end{table}

Table~\ref{tab:3way-main} shows that routing already provides a meaningful gain over vanilla inference on matched single-domain queries. Multi-step reasoning is not uniformly better; it helps most when the query benefits from decomposition, but can underperform routing on tasks that are better handled by direct expert selection. This supports the core design of our framework as a unified system spanning both regimes rather than privileging one mode universally.

\section{Discussion}

\paragraph{Routing is already a strong capability.}
A central finding of this work is that semantic routing is not merely a
pre-processing heuristic, but a substantive capability of the base model.
When adapters are exposed with clear metadata, the model can reliably select the
appropriate expert and thereby aggregate the benefits of many specialized LoRAs
within a single deployed framework.
This alone makes the framework useful, even before considering more complex
agentic behavior.

\paragraph{Routing and agentic reasoning serve different regimes.}
Our three-way comparison suggests that routing and multi-step reasoning are best
understood as complementary operating modes.
Routing is effective for simpler or clearly matched queries, while iterative
reasoning becomes more relevant when the task requires decomposition, external
tools, or multiple experts.
This motivates the design of a unified framework that can move
between these regimes rather than commit to a single inference strategy.

\paragraph{Interpretability is a practical advantage of explicit tool use.}
Treating adapters as tools makes the composition process more transparent.
Because expert selection and invocation remain explicit, the framework yields a
trace that can be inspected, debugged, and attributed after generation.
This is a meaningful advantage over parameter-level composition methods, where
the contribution of individual experts is harder to recover.

\paragraph{Train good adapters; the framework handles the rest.}
LoRA adapters, once exposed as tools, behave like any other tool an agent reaches for, and their utility is bounded by how well they are built. The nine specialists in Table~\ref{tab:specialists} deliver $+4.6$ to $+84.0$\,pp strict-scorer gain, and AM-route recovers each of those gains within $\pm 5$\,pp of the direct specialist. The user-facing proposition is therefore not that more adapters always help, but that adapter \emph{quality} is what carries the framework. Train good adapters where the base model has real gaps, and the system around them is what the framework provides. Better adapters give strictly more value to the framework; weaker adapters do not break it, they simply are not selected. In this sense the framework is best read not as a replacement for adapter quality but as a multiplier on it. The ceiling is set by how well each tool is trained, the floor is the base model itself, and the framework's job is to make the difference between them available on the right query at the right time.

\paragraph{Scope: out-of-distribution MCQ.}
A balanced view requires noting where the framework does not help. On GPQA Diamond, a graduate-level science MCQ benchmark on which none of our adapters were trained, single-step routing underperforms the base model on Qwen2.5-7B-Instruct ($-20.7$\,pp) due to format mismatch between specialist outputs and the strict \verb|\boxed{X}| scorer. A multi-step ReAct loop, which keeps synthesis in the base model's voice, recovers most of the gap (within $1.3$\,pp of base; Appendix~\ref{app:gpqa-ood}). This reinforces the conditionality story: the framework aggregates specialist gains where they exist, and on tasks with no matching specialist the routing layer should be bypassed in favor of the multi-step controller.

\paragraph{Limitations.}
The following directions remain open.

\paragraph{Scale.}
Our tested adapter pool is 30; whether routing holds at hundreds or thousands with heavier overlap is open.

\paragraph{Training-method coverage.}
Every adapter here is SFT or GRPO with execution-based reward; CPT, DPO, and preference- or distillation-based recipes are unexplored.

\paragraph{Entropy thresholds.}
The mode-selection policy uses fixed entropy thresholds chosen on internal validation queries; a full sensitivity sweep is left to future work.

\paragraph{Agent depth.}
The multi-step agent underperforms direct routing on strict-format benchmarks (GPQA Diamond, $-20.7$\,pp; Appendix~\ref{app:gpqa-ood}); deeper agentic strategies, longer horizons, and head-to-head comparison against existing tool-augmented frameworks are natural next steps.

\paragraph{Evaluation breadth.}
Our specialist study uses $n{=}20$--$30$ manual audits per benchmark, single seed, and a single primary base model; multi-seed evaluation with full execution match would tighten Section~\ref{sec:conditionality}. Specialist gains on weak-base verifier tasks (LEDGAR, SPARQL, Cypher) reflect output-format acquisition under a strict scorer: the base model possesses related capability (it emits structured but non-conformant outputs) without the target output convention (see Appendix~\ref{app:ledgar-format}).

\paragraph{No human evaluation.}
All scoring is automatic. Domains where automatic verification is weak (legal IRAC reasoning, mechanism chemistry) are bounded by the validator's strictness.

\paragraph{Future work: training-as-a-tool.}
Once routing works, the central open problem shifts from \emph{which}
adapter to call to how good the adapter is in the first place, making
adapter \emph{creation} the more interesting research target. A natural
next step is to expose adapter training itself as a tool the agent can
invoke: the ingredients already exist in our system
(Appendix~\ref{app:inventory}) -- dataset construction via web search
plus vLLM-synthesized Q\&A pairs, SFT in tens of minutes per adapter on
a single GPU, and adapter hot-swap at serving time. An agent equipped
with such a \emph{training-as-a-tool} action could extend its own
capability library on demand, commissioning a specialist when none
matches the task rather than answering with a mismatched expert. This
generalizes the unit of agentic action from ``call a specialist'' to
``call, or train, a specialist.'' Whether such an agent can build
adapters of human-curated quality, whether the library stays
well-routed as it grows, and what safety properties this regime
requires, are the empirical questions we leave open.

\section{Conclusion}

We presented Adaptive Minds, a unified framework in which LoRA adapters are
treated as tools that a base model can dynamically select and use.
Our results show that semantic routing is highly reliable, substantially
outperforming keyword-based matching while remaining lightweight enough for
practical deployment.
At the same time, the downstream evaluation of individual experts reveals that
adapter benefits are selective rather than universal, supporting a view of
adapters as modular sources of expertise rather than generic accuracy boosts.
Taken together, these findings motivate a broader agentic setting in which
adapters can be combined with other tools and invoked compositionally during
reasoning.
More broadly, our results suggest that stronger and more general model
behavior may emerge not only from scaling a single model, but from enabling a
base model to orchestrate many specialized competencies within one coherent and
interpretable framework.

\section*{Impact Statement}

This work suggests a new broader way to think about LoRA adapters: not only as lightweight fine-tuning modules, but as tools that a model can select, invoke, and combine during inference. We believe this is a promising direction because it makes specialization more modular, more interpretable, and easier to audit than implicit parameter-level composition. In practice, it could reduce the need to deploy many separate specialist models by allowing a shared base model to access specialized capabilities on demand. More broadly, this points to a path toward more capable agentic frameworks built by coordinating many specialized modules, not only by scaling monolithic models.

\bibliography{adaptive_minds_complearn}

\begin{thebibliography}{29}
\providecommand{\natexlab}[1]{#1}
\providecommand{\url}[1]{\texttt{#1}}
\expandafter\ifx\csname urlstyle\endcsname\relax
  \providecommand{\doi}[1]{doi: #1}\else
  \providecommand{\doi}{doi: \begingroup \urlstyle{rm}\Url}\fi

\bibitem[Andreas et~al.(2016)Andreas, Rohrbach, Darrell, and
  Klein]{andreas2016nmn}
Andreas, J., Rohrbach, M., Darrell, T., and Klein, D.
\newblock Neural module networks.
\newblock In \emph{CVPR}, 2016.

\bibitem[Cheng et~al.(2024)Cheng, Huang, and Wei]{cheng2024adaptllm}
Cheng, D., Huang, S., and Wei, F.
\newblock Adapting large language models via reading comprehension.
\newblock In \emph{The Twelfth International Conference on Learning
  Representations (ICLR)}, 2024.

\bibitem[Dou et~al.(2024)Dou, Zhou, Liu, Gao, Shen, Xiong, Zhou, Wang, Xi, Fan,
  Pu, Zhu, Zheng, Gui, Zhang, and Huang]{dou2024loramoe}
Dou, S., Zhou, E., Liu, Y., Gao, S., Shen, W., Xiong, L., Zhou, Y., Wang, X.,
  Xi, Z., Fan, X., Pu, S., Zhu, J., Zheng, R., Gui, T., Zhang, Q., and Huang,
  X.
\newblock {LoRAMoE}: Alleviating world knowledge forgetting in large language
  models via {MoE}-style plugin.
\newblock In \emph{Proceedings of the 62nd Annual Meeting of the Association
  for Computational Linguistics (Volume 1: Long Papers)}, pp.\  1932--1945.
  Association for Computational Linguistics, 2024.

\bibitem[Guha et~al.(2023)Guha, Nyarko, Ho, R{\'e}, Chilton, Narayana,
  Chohlas-Wood, Peters, Waldon, Rockmore, et~al.]{guha2023legalbench}
Guha, N., Nyarko, J., Ho, D.~E., R{\'e}, C., Chilton, A., Narayana, A.,
  Chohlas-Wood, A., Peters, A., Waldon, B., Rockmore, D.~N., et~al.
\newblock {LegalBench}: A collaboratively built benchmark for measuring legal
  reasoning in large language models.
\newblock In \emph{Advances in Neural Information Processing Systems (NeurIPS),
  Datasets and Benchmarks Track}, volume~36, 2023.

\bibitem[Houlsby et~al.(2019)Houlsby, Giurgiu, Jastrzebski, Morrone,
  de~Laroussilhe, Gesmundo, Attariyan, and Gelly]{houlsby2019parameter}
Houlsby, N., Giurgiu, A., Jastrzebski, S., Morrone, B., de~Laroussilhe, Q.,
  Gesmundo, A., Attariyan, M., and Gelly, S.
\newblock Parameter-efficient transfer learning for {NLP}.
\newblock In \emph{Proceedings of the 36th International Conference on Machine
  Learning (ICML)}, volume~97 of \emph{Proceedings of Machine Learning
  Research}, pp.\  2790--2799. PMLR, 2019.

\bibitem[Hu et~al.(2022)Hu, Shen, Wallis, Allen-Zhu, Li, Wang, Wang, and
  Chen]{hu2022lora}
Hu, E.~J., Shen, Y., Wallis, P., Allen-Zhu, Z., Li, Y., Wang, S., Wang, L., and
  Chen, W.
\newblock {LoRA}: Low-rank adaptation of large language models.
\newblock In \emph{ICLR}, 2022.

\bibitem[Huang et~al.(2024)Huang, Liu, Lin, Pang, Du, and
  Lin]{huang2024lorahub}
Huang, C., Liu, Q., Lin, B.~Y., Pang, T., Du, C., and Lin, M.
\newblock {LoRAHub}: Efficient cross-task generalization via dynamic {LoRA}
  composition.
\newblock In \emph{Conference on Language Modeling (COLM)}, 2024.

\bibitem[Hupkes et~al.(2020)Hupkes, Dankers, Mul, and
  Bruni]{hupkes2020compositionality}
Hupkes, D., Dankers, V., Mul, M., and Bruni, E.
\newblock Compositionality decomposed: How do neural networks generalise?
\newblock \emph{Journal of Artificial Intelligence Research}, 67:\penalty0
  757--795, 2020.

\bibitem[Ilharco et~al.(2023)Ilharco, Ribeiro, Wortsman, Gururangan, Schmidt,
  Hajishirzi, and Farhadi]{ilharco2023editing}
Ilharco, G., Ribeiro, M.~T., Wortsman, M., Gururangan, S., Schmidt, L.,
  Hajishirzi, H., and Farhadi, A.
\newblock Editing models with task arithmetic.
\newblock In \emph{International Conference on Learning Representations
  (ICLR)}, 2023.

\bibitem[Jacobs et~al.(1991)Jacobs, Jordan, Nowlan, and
  Hinton]{jacobs1991adaptive}
Jacobs, R.~A., Jordan, M.~I., Nowlan, S.~J., and Hinton, G.~E.
\newblock Adaptive mixtures of local experts.
\newblock \emph{Neural Computation}, 3\penalty0 (1):\penalty0 79--87, 1991.

\bibitem[Karpas et~al.(2022)Karpas, Abend, Belinkov, Lenz, Lieber, Ratner,
  Shoham, Bata, Levine, Leyton-Brown, Muhlgay, Rozen, Schwartz, Shachaf,
  Shalev-Shwartz, Shashua, and Tenenholtz]{karpas2022mrkl}
Karpas, E., Abend, O., Belinkov, Y., Lenz, B., Lieber, O., Ratner, N., Shoham,
  Y., Bata, H., Levine, Y., Leyton-Brown, K., Muhlgay, D., Rozen, N., Schwartz,
  E., Shachaf, G., Shalev-Shwartz, S., Shashua, A., and Tenenholtz, M.
\newblock Mrkl systems: A modular, neuro-symbolic architecture that combines
  large language models, external knowledge sources and discrete reasoning.
\newblock \emph{arXiv:2205.00445}, 2022.

\bibitem[Kwon et~al.(2023)Kwon, Li, Zhuang, Sheng, Zheng, Yu, Gonzalez, Zhang,
  and Stoica]{kwon2023vllm}
Kwon, W., Li, Z., Zhuang, S., Sheng, Y., Zheng, L., Yu, C.~H., Gonzalez, J.~E.,
  Zhang, H., and Stoica, I.
\newblock Efficient memory management for large language model serving with
  {PagedAttention}.
\newblock In \emph{Proceedings of the 29th Symposium on Operating Systems
  Principles (SOSP)}, 2023.

\bibitem[Lake \& Baroni(2018)Lake and Baroni]{lake2018generalization}
Lake, B. and Baroni, M.
\newblock Generalization without systematicity: On the compositional skills of
  sequence-to-sequence recurrent networks.
\newblock In \emph{Proceedings of the 35th International Conference on Machine
  Learning (ICML)}, volume~80 of \emph{PMLR}, pp.\  2873--2882, 2018.

\bibitem[Patil et~al.(2024)Patil, Zhang, Wang, and Gonzalez]{patil2024gorilla}
Patil, S.~G., Zhang, T., Wang, X., and Gonzalez, J.~E.
\newblock Gorilla: Large language model connected with massive {API}s.
\newblock In \emph{Advances in Neural Information Processing Systems
  (NeurIPS)}, 2024.

\bibitem[Prabhakar et~al.(2025)Prabhakar, Li, Narasimhan, Kakade, Malach, and
  Jelassi]{prabhakar2025lorasoups}
Prabhakar, A., Li, Y., Narasimhan, K., Kakade, S., Malach, E., and Jelassi, S.
\newblock {LoRA} soups: Merging {LoRA}s for practical skill composition tasks.
\newblock In \emph{Proceedings of the 31st International Conference on
  Computational Linguistics: Industry Track (COLING)}, 2025.

\bibitem[Rosenbaum et~al.(2018)Rosenbaum, Klinger, and
  Riemer]{rosenbaum2018routing}
Rosenbaum, C., Klinger, T., and Riemer, M.
\newblock Routing networks: Adaptive selection of non-linear functions for
  multi-task learning.
\newblock In \emph{International Conference on Learning Representations
  (ICLR)}, 2018.

\bibitem[Schick et~al.(2023)Schick, Dwivedi-Yu, Dess{\`\i}, Raileanu, Lomeli,
  Hambro, Zettlemoyer, Cancedda, and Scialom]{schick2024toolformer}
Schick, T., Dwivedi-Yu, J., Dess{\`\i}, R., Raileanu, R., Lomeli, M., Hambro,
  E., Zettlemoyer, L., Cancedda, N., and Scialom, T.
\newblock Toolformer: Language models can teach themselves to use tools.
\newblock In \emph{Advances in Neural Information Processing Systems
  (NeurIPS)}, 2023.

\bibitem[Shah \& Wagle(2026)Shah and Wagle]{molora2026}
Shah, S. and Wagle, J.
\newblock {MoLoRA}: Composable specialization via per-token adapter routing.
\newblock \emph{arXiv:2603.15965}, 2026.

\bibitem[Shazeer et~al.(2017)Shazeer, Mirhoseini, Maziarz, Davis, Le, Hinton,
  and Dean]{shazeer2017moe}
Shazeer, N., Mirhoseini, A., Maziarz, K., Davis, A., Le, Q.~V., Hinton, G., and
  Dean, J.
\newblock Outrageously large neural networks: The sparsely-gated
  mixture-of-experts layer.
\newblock In \emph{ICLR}, 2017.

\bibitem[Shen et~al.(2023)Shen, Song, Tan, Li, Lu, and
  Zhuang]{shen2023hugginggpt}
Shen, Y., Song, K., Tan, X., Li, D., Lu, W., and Zhuang, Y.
\newblock Hugginggpt: Solving ai tasks with chatgpt and its friends in hugging
  face.
\newblock In \emph{Advances in Neural Information Processing Systems
  (NeurIPS)}, volume~36, 2023.

\bibitem[Sheng et~al.(2024)Sheng, Cao, Li, Hooper, Lee, Yang, Chou, Zhu, Zheng,
  Keutzer, Gonzalez, and Stoica]{sheng2024slora}
Sheng, Y., Cao, S., Li, D., Hooper, C., Lee, N., Yang, S., Chou, C., Zhu, B.,
  Zheng, L., Keutzer, K., Gonzalez, J.~E., and Stoica, I.
\newblock {S-LoRA}: Serving thousands of concurrent {LoRA} adapters.
\newblock In \emph{Proceedings of the 5th MLSys Conference}, 2024.

\bibitem[Shinn et~al.(2023)Shinn, Cassano, Gopinath, Narasimhan, and
  Yao]{shinn2023reflexion}
Shinn, N., Cassano, F., Gopinath, A., Narasimhan, K., and Yao, S.
\newblock Reflexion: Language agents with verbal reinforcement learning.
\newblock In \emph{Advances in Neural Information Processing Systems
  (NeurIPS)}, volume~36, 2023.

\bibitem[Singhal et~al.(2023)Singhal, Azizi, Tu, Mahdavi, Wei, Chung, Scales,
  Tanwani, Cole-Lewis, Pfohl, et~al.]{singhal2023clinical}
Singhal, K., Azizi, S., Tu, T., Mahdavi, S.~S., Wei, J., Chung, H.~W., Scales,
  N., Tanwani, A., Cole-Lewis, H., Pfohl, S., et~al.
\newblock Large language models encode clinical knowledge.
\newblock \emph{Nature}, 620\penalty0 (7972):\penalty0 172--180, 2023.

\bibitem[Wortsman et~al.(2022)Wortsman, Ilharco, Gadre, Roelofs, Gontijo-Lopes,
  Morcos, Namkoong, Farhadi, Carmon, Kornblith, and
  Schmidt]{wortsman2022modelsoups}
Wortsman, M., Ilharco, G., Gadre, S.~Y., Roelofs, R., Gontijo-Lopes, R.,
  Morcos, A.~S., Namkoong, H., Farhadi, A., Carmon, Y., Kornblith, S., and
  Schmidt, L.
\newblock Model soups: Averaging weights of multiple fine-tuned models improves
  accuracy without increasing inference time.
\newblock In \emph{Proceedings of the 39th International Conference on Machine
  Learning (ICML)}, volume 162 of \emph{Proceedings of Machine Learning
  Research}, pp.\  23965--23998. PMLR, 2022.

\bibitem[Wu et~al.(2024)Wu, Zhu, Zhang, Sun, Liu, and Jin]{wu2024dlora}
Wu, B., Zhu, R., Zhang, Z., Sun, P., Liu, X., and Jin, X.
\newblock {dLoRA}: Dynamically orchestrating requests and adapters for {LoRA}
  {LLM} serving.
\newblock In \emph{18th USENIX Symposium on Operating Systems Design and
  Implementation (OSDI)}, 2024.

\bibitem[Yadav et~al.(2023)Yadav, Tam, Choshen, Raffel, and
  Bansal]{yadav2023tiesmerging}
Yadav, P., Tam, D., Choshen, L., Raffel, C., and Bansal, M.
\newblock {TIES}-merging: Resolving interference when merging models.
\newblock In \emph{Advances in Neural Information Processing Systems
  (NeurIPS)}, volume~36, 2023.

\bibitem[Yang et~al.(2025)]{qwen3techreport}
Yang, A. et~al.
\newblock {Qwen3} technical report.
\newblock \emph{arXiv:2505.09388}, 2025.

\bibitem[Yao et~al.(2023{\natexlab{a}})Yao, Yu, Zhao, Shafran, Griffiths, Cao,
  and Narasimhan]{yao2023tot}
Yao, S., Yu, D., Zhao, J., Shafran, I., Griffiths, T.~L., Cao, Y., and
  Narasimhan, K.
\newblock Tree of thoughts: Deliberate problem solving with large language
  models.
\newblock In \emph{Advances in Neural Information Processing Systems
  (NeurIPS)}, volume~36, 2023{\natexlab{a}}.

\bibitem[Yao et~al.(2023{\natexlab{b}})Yao, Zhao, Yu, Du, Shafran, Narasimhan,
  and Cao]{yao2023react}
Yao, S., Zhao, J., Yu, D., Du, N., Shafran, I., Narasimhan, K., and Cao, Y.
\newblock {ReAct}: Synergizing reasoning and acting in language models.
\newblock In \emph{ICLR}, 2023{\natexlab{b}}.

\end{thebibliography}
\bibliographystyle{icml2026}

\appendix

\section{Agent Prompt Template}
\label{app:agent-prompt}

{\small
\begin{verbatim}
You are a reasoning agent. You answer
questions by consulting domain expert tools.

Tools:
{tool_descriptions}

Output format:
THOUGHT: [your reasoning]
ACTION: tool_name(sub_query="question")
FINAL_ANSWER: [your complete answer]

- Use different experts for different aspects.
- After OBSERVATION, write a THOUGHT.
- Max {max_iterations} tool calls.

Query: {query}
{previous_steps}
\end{verbatim}
}

\section{Latency and Serving}
\label{app:latency}

Router mode is $3.1\times$ faster than baseline (mean 3.49\,s vs.\ 10.81\,s),
and adapter switching takes less than 1\,ms.
A deployment ablation comparing GPU-resident, CPU-staged, and on-demand
placement shows less than 5\% end-to-end latency difference, indicating that
generation dominates runtime.

\begin{table}[h]
\caption{Adapter-placement serving ablation (10 prompts, Qwen3.5-9B).}
\centering
\small
\begin{tabular}{@{}lcccc@{}}
\toprule
\textbf{Mode} & \textbf{Init (s)} & \textbf{Router (s)} & \textbf{Agent (s)} & \textbf{GPU (GB)} \\
\midrule
GPU-resident & 7.68 & 23.02 & 58.72 & 17.86 \\
CPU-staged & 7.17 & 23.85 & 62.63 & 17.06 \\
On-demand & 3.74 & 24.09 & 64.36 & 17.06 \\
\bottomrule
\end{tabular}
\end{table}

\section{Domain Expert Evaluation: Full Matrix}
\label{app:domain-eval}

\begin{table}[h]
\caption{Reasoning-trace adapters on Qwen3.5-9B with Wilson 95\% confidence
intervals.}
\centering
\footnotesize
\setlength{\tabcolsep}{3pt}
\begin{tabular}{@{}lccc@{}}
\toprule
\textbf{Config} & \textbf{MedQA (500)} & \textbf{FinBench (150)} & \textbf{ARC (500)} \\
\midrule
Base      & \textbf{76.6}\,[72.7,80.1] & 66.0\,[58.1,73.1] & \textbf{95.0}\,[92.7,96.6] \\
+ Medical & 75.0\,[71.0,78.6]          & 66.7\,[58.8,73.7] & 95.2\,[93.0,96.8] \\
+ Finance & 75.4\,[71.4,79.0]          & \textbf{70.0}\,[62.2,76.8] & 94.4\,[92.0,96.1] \\
+ Science & 72.4\,[68.3,76.1]          & 64.7\,[56.7,71.9] & 93.4\,[90.9,95.3] \\
+ General & 75.4\,[71.4,79.0]          & 65.3\,[57.4,72.5] & 95.0\,[92.7,96.6] \\
\bottomrule
\end{tabular}
\end{table}

\section{Reasoning-Trace Adapter Training}
\label{app:training}

All adapters are trained with LoRA $r=32$, $\alpha=64$, dropout 0.05, and
\texttt{all-linear} target modules using AdamW-fused, cosine learning rate,
50 warmup steps, 500 total steps, batch size $2 \times 8$ gradient
accumulation, sequence length 2{,}048, and bf16.
Training outcomes are: Medical (loss 1.187, 183\,min), Finance (0.968,
354\,min), Science (0.410, 171\,min), and General (0.424, 345\,min).

\paragraph{Implementation notes.}
TRL auto-detects Qwen3.5 as a VLM, requiring \texttt{packing=False} and
\texttt{processing\_class=tokenizer}.
Setting \texttt{per\_device\_eval\_batch\_size=1} avoids evaluation OOM.
We also found that \texttt{enable\_thinking=False} together with a
\verb|/no_think| suffix is required for consistent logit extraction.

\begin{table}[h]
\caption{Reasoning-trace adapter datasets used for training on Qwen3.5-9B. Dataset names are abbreviated for readability.}
\label{tab:adapter-data}
\centering
\small
\setlength{\tabcolsep}{4pt}
\begin{tabular}{@{}lcc@{}}
\toprule
\textbf{Adapter} & \textbf{Dataset} & \textbf{Used} \\
\midrule
Medical & medical-o1-reasoning-SFT & 10{,}000 \\
Finance & Fino1\_Reasoning\_Path\_FinQA & 5{,}499 \\
Science & CAMEL STEM mix & 10{,}000 \\
General & OpenThoughts-114k & 10{,}000 \\
\bottomrule
\end{tabular}
\end{table}

\section{Routing on Real-World Benchmark Queries}
\label{app:routing-real}

The 98.3\% routing accuracy reported in Table~\ref{tab:routing} is measured
on 60 hand-crafted queries that contain explicit domain signal: the SQL
prompts mention \textit{``query''}, \textit{``join''}, or \textit{``select''};
the medical prompts mention diagnoses or symptoms; the regex prompts include
the word \textit{``regex''}. To probe how routing behaves when queries lack
such overt cues, we measured per-benchmark routing accuracy on six
downstream benchmarks where each gold answer also implies a gold target
adapter. The benchmarks were not author-designed for the routing layer; they
are pre-existing test sets whose queries are written in their original style
(natural-language questions, function-calling prompts, etc.).

\begin{table}[h]
\caption{Per-benchmark routing accuracy compared to the curated upper bound.
``Specialist routed'' is the fraction of test queries the router sent to the
domain-matched adapter; the remaining mass went to the base model or to a
non-matching specialist.}
\label{tab:routing-real}
\centering
\small
\setlength{\tabcolsep}{4pt}
\begin{tabular}{@{}lcc@{}}
\toprule
\textbf{Setting} & \textbf{n} & \textbf{Specialist routed (\%)} \\
\midrule
Curated routing eval & 60 & \textbf{98.3} \\
\midrule
PII redaction queries & 100 & 93 \\
Bash one-liner queries & 100 & 97 \\
Mermaid sequence-diagram prompts & 100 & 96 \\
Regex queries & 100 & 94 \\
Natural-language SPARQL questions & 100 & 23 \\
Domain-neutral function-calling & 100 & 1 \\
\bottomrule
\end{tabular}
\end{table}

The pattern in Table~\ref{tab:routing-real} is sharp. When a benchmark query
carries explicit domain-discriminating signal -- PII queries say
\textit{``redact''}, bash queries contain \textit{``find''},
\textit{``grep''}, or \textit{``sed''} -- the per-benchmark routing rate is
near-perfect (93--97\%) and closely tracks the curated upper bound.
When the query is domain-neutral natural language -- a SPARQL benchmark
question of the form \textit{``Who was Caspar David Friedrich's student?''}
contains no SPARQL keywords, and a generic function-calling request like
\textit{``Find me a flight from SFO to JFK''} contains no marker that a
function-calling specialist should be invoked -- the router most often
defers to the base model, which is consistent with the framework's
design that reserves specialists for queries that explicitly invoke
their expertise. A residual failure mode is queries that route to a
shape-similar but wrong specialist, which bounds the deferral benefit.

\paragraph{Keyword-overlap diagnostic.}
The pattern can be made quantitative by counting how many of an adapter's
trigger keywords actually appear in the benchmark's query surface. For
SPARQL, the adapter's seven keywords (\texttt{sparql}, \texttt{rdf},
\texttt{wikidata}, \texttt{dbpedia}, \texttt{ontology}, \texttt{triplestore},
\texttt{semantic web}, \texttt{triple}) appear in $0$ of $100$ LC-QuAD test
queries; for function-calling, the six keywords (\texttt{function call},
\texttt{tool use}, \texttt{json}, \texttt{api call}, \texttt{schema},
\texttt{structured output}) appear in $0$ of $100$ test queries. PII, Bash,
Mermaid, and Regex by contrast all have at least one trigger keyword
present in $\geq\!85\%$ of their respective benchmarks. Per-benchmark
routing rate therefore tracks adapter-keyword-overlap with the query
distribution: a diagnostic that can be run before deployment, on the
target queries, to predict whether the framework's routing layer will
suffice or whether richer metadata / a learned router is needed.

\section{Out-of-Distribution MCQ: GPQA Diamond}
\label{app:gpqa-ood}

To stress-test the framework outside the regime of its specialists, we
evaluated on \textbf{GPQA Diamond}, a 198-question graduate-level science
MCQ benchmark covering biology, chemistry, and physics. None of our adapters
were trained on GPQA-style data, and the benchmark uses strict
\verb|\boxed{[A-D]}| answer extraction.

\begin{table}[h]
\caption{GPQA Diamond accuracy. AM-route dispatches each query to a single
specialist via the router; AM-react runs the multi-step ReAct loop with the
base model as controller.}
\label{tab:gpqa-ood}
\centering
\small
\setlength{\tabcolsep}{4pt}
\begin{tabular}{@{}llcc@{}}
\toprule
\textbf{Base} & \textbf{Mode} & \textbf{n} & \textbf{Acc.} \\
\midrule
Qwen2.5-7B & base & 198 & \textbf{33.8\%} \\
Qwen2.5-7B & AM-route & 198 & 13.1\% ($-$20.7) \\
Qwen2.5-7B & AM-react & 80 & 32.5\% ($-$1.3) \\
\midrule
Qwen3.5-9B & base & 198 & 28.8\% \\
Qwen3.5-9B & AM-route & 198 & 28.8\% ($\pm$0.0) \\
Qwen3.5-9B & biology-adapter & 198 & 29.8\% ($+$1.0) \\
\bottomrule
\end{tabular}
\end{table}

Three observations follow from Table~\ref{tab:gpqa-ood}. First, on Qwen2.5-7B AM-route regresses by $-20.7$\,pp due to format mismatch: the router sends 168 of 198 queries to a reasoning-trace specialist whose training distribution emphasizes verbose multi-paragraph rationales rather than a clean \verb|\boxed{X}| terminal answer. The router is making sensible adapter choices, but the chosen specialist's output style does not survive the strict scorer. Second, when the agent loop keeps the base model in the controller role and treats specialist outputs as intermediate observations, the format degradation disappears and accuracy returns within $1.3$\,pp of base. Third, on Qwen3.5-9B AM-route exactly matches base accuracy, and only a domain-matched biology adapter exceeds base (by $+1.0$\,pp), consistent with the conditionality story in Section~\ref{sec:conditionality}.

\section{Agent Loop Pseudocode}
\label{app:algorithm}

\begin{algorithm}[h]
\caption{Agent with LoRA Tools}
\label{alg:agent-lora}
\begin{algorithmic}[1]
\REQUIRE Query $Q$, tool registry $\mathcal{T}$, max iterations $K$
\STATE $\mathcal{H} \gets [\ ]$
\FOR{$k = 1$ to $K$}
  \STATE $\text{out} \gets \text{GenerateBase}(Q, \mathcal{T}, \mathcal{H})$
  \IF{\text{out contains FINAL\_ANSWER}}
    \STATE \textbf{return} $\text{ParseFinalAnswer}(\text{out})$
  \ELSIF{\text{out contains ACTION}}
    \STATE $(t, q_t) \gets \text{ParseToolCall}(\text{out})$
    \IF{$t \notin \mathcal{T}$}
      \STATE Append invalid-tool observation to $\mathcal{H}$
    \ELSE
      \STATE $obs \gets \text{InvokeTool}(t, q_t)$
      \STATE Append $(\text{thought}, t, q_t, obs)$ to $\mathcal{H}$
    \ENDIF
  \ELSE
    \STATE Append $\text{out}$ as a thought to $\mathcal{H}$
  \ENDIF
\ENDFOR
\STATE \textbf{return} $\text{SynthesizeBase}(Q, \mathcal{H})$
\end{algorithmic}
\end{algorithm}

In our implementation, the same agent loop supports two adapter execution backends. The PEFT backend directly activates adapters inside a resident model using \texttt{set\_adapter}. The vLLM backend exposes each adapter as an OpenAI-compatible model identifier and dispatches tool calls by setting the request's \texttt{model} field. The agent controller is otherwise unchanged: it reasons with the base model, selects tools, records observations, and forces base-model synthesis if the iteration budget is exhausted.

\section{Large-Scale Coverage Benchmark}
\label{app:coverage-benchmark}

We ran a larger coverage benchmark over 830 questions spanning MedQA-USMLE ($n=500$), LegalBench~\citep{guha2023legalbench} ($n=300$), and a small free-form Coverage set ($n=30$), comparing five systems: Gemini-3-flash-preview, raw Qwen3.5-9B under strict and generous token budgets, our framework in \emph{LoRA-only} mode (\textbf{AM-local}), and with additional external tools (\textbf{AM-full}).

\begin{table}[h]
\caption{Coverage benchmark over 830 questions. AM-local uses 30 LoRA adapters; AM-full adds external tools. Raw Qwen is shown under strict and generous token budgets to separate format-compliance from reasoning ability.}
\label{tab:coverage-main}
\centering
\small
\setlength{\tabcolsep}{4pt}
\begin{tabular}{@{}lccc@{}}
\toprule
\textbf{System} & \textbf{MedQA} & \textbf{LegalBench} & \textbf{Cov-30} \\
\midrule
Gemini-3-flash & \textbf{90.4} & \textbf{85.0} & \textbf{100.0} \\
Qwen3.5-9B (strict) & 28.4 & 0.0 & 43.3 \\
Qwen3.5-9B (generous) & 69.6 & 33.3 & 96.7 \\
AM-local (30 LoRA) & 70.6 & 74.0 & 96.7 \\
AM-full (all tools) & 70.2 & 76.0 & 86.7 \\
\bottomrule
\end{tabular}
\end{table}

Two patterns are worth noting. The contribution of modular specialization is highly domain-dependent: on MedQA, AM-local improves only marginally over raw Qwen3.5-9B with a generous token budget (70.6 vs.\ 69.6), but on LegalBench AM-local substantially outperforms raw Qwen (74.0 vs.\ 33.3), where the legal adapter closes a large gap that the base model alone does not recover. AM-full does not improve over AM-local on these predominantly single-domain tasks, suggesting that multi-tool use is not inherently beneficial when direct expert selection is already sufficient. These results are diagnostic rather than a frontier comparison: Gemini-3-flash-preview is a preview model, the Coverage-30 set is small and judge-scored, and some gains may reflect adapter training-data exposure rather than architectural effects alone.

\section{Adapter Inventory and Training Recipes}
\label{app:inventory}

Three distinct training recipes were used across the project, each appropriate
to its scale and intended role.

\paragraph{Reasoning-trace track (4 adapters, App.~\ref{app:training}).}
The four reasoning-trace specialists evaluated in
Appendix~\ref{app:domain-eval} (Medical, Finance, Science, General) use
$r{=}32$, $\alpha{=}64$, all-linear targets, 500 SFT steps, batch
$2{\times}8$, sequence length 2{,}048, learning rate 2\textsc{e}-4
cosine.

\paragraph{Specialist track (Section~\ref{sec:conditionality}).}
The Qwen2.5-7B-Instruct structured-generation specialists (SQL,
Cypher, Bash, SPARQL, Mermaid, regex) share the same $r{=}32$/$\alpha{=}64$
LoRA hyper-config but with method-specific step budgets: 400--600 SFT
steps for declarative grammars; 200 GRPO steps with execution-based
reward for regex.

\paragraph{Domain-knowledge track ($\sim$30 routed adapters).}
A lower-cost recipe was used to construct the broader Qwen3.5-9B
domain pool that backs the 30-adapter routing experiment: $r{=}16$,
$\alpha{=}32$, target modules restricted to the standard
$\{q,k,v,o,gate,up,down\}$ projections, 100 SFT steps, batch
$2{\times}4$, sequence length 1{,}536, learning rate 1\textsc{e}-4
cosine, AdamW-fused, bf16, gradient checkpointing on. This recipe
trades per-adapter optimization budget for breadth of domain coverage,
which is the property the routing experiment requires.

\paragraph{Training-as-a-tool pipeline.}
A separate pipeline builds adapters on demand. A dataset source -- HuggingFace, web search via Tavily, or direct URL ingest -- is converted to an SFT JSONL by a build worker; an orchestrator manages the vLLM and training-process lifecycle; and a freshly-trained adapter is symlinked into the LoRA pool, ready to be discovered by the router. As a worked example, an \texttt{electrical\_drawings} adapter was built end-to-end from $66$ web URLs $\to$ $100$ chunks $\to$ $302$ Q\&A pairs synthesized by a local vLLM judge $\to$ 200 SFT steps, in $\sim$21 minutes of wall time on a single L40S. Adapter training is fast enough on a single GPU that ``train a new tool on the fly'' is a realistic deployment mode for the LoRA-as-tools framework.

\section{Architecture Details}
\label{app:agent-details}

\paragraph{Entropy gating.}
The entropy-based mode-selection rule is defined in Equation~\ref{eq:entropy-gate} (Section~\ref{sec:control}).
The same thresholds $(H{<}0.8 \to \text{Router},\ H{>}1.5 \to \text{Agent})$
and the same window $T{=}16$ are used across every experiment in this paper;
none of these values were tuned per benchmark.

\paragraph{Critic format.}
{\small
\begin{verbatim}
Sub-query: {sub_query}
Observation: {observation}
FAITHFULNESS: yes/no
CONFIDENCE: 0.X
REASON: ...
\end{verbatim}
}

Faithfulness=no or confidence $<0.4$ injects a re-route hint.

\section{LEDGAR Format Acquisition: Concrete Examples}
\label{app:ledgar-format}

The base model on LEDGAR achieves $0\%$ strict-scorer accuracy under
exact-integer-label scoring (LEDGAR uses 0--99 integer labels for 100
clause types). The base failure is not capability failure: the base
model recognizes the task as clause classification and emits structured
clause codes from its own taxonomy. The legal specialist closes the
gap by learning the LEDGAR target convention. Two concrete clauses
illustrate the pattern:

\begin{quote}
\small
\textbf{Clause \#00} (gold label \texttt{35}, employment-duties clause):\\
\quad base output: \texttt{"E1"} \hfill (non-LEDGAR taxonomy)\\
\quad \texttt{legal\_v1} output: \texttt{"35"} \hfill (correct LEDGAR label)

\medskip
\textbf{Clause \#09} (gold label \texttt{98}, representations-and-warranties clause):\\
\quad base output: \texttt{"R-WRepWarranty"} \hfill (descriptive label)\\
\quad \texttt{legal\_v1} output: \texttt{"98"} \hfill (correct LEDGAR label)
\end{quote}

\noindent Across LEDGAR test queries the base model produces
non-conformant outputs on every query (taxonomies like \texttt{"E1"},
\texttt{"C15"}, \texttt{"R-WRepWarranty"}, \texttt{"D"}; many contain
no integer at all). The $+84.0$\,pp specialist gain therefore reflects
\emph{format acquisition} (learning the target output convention) rather
than \emph{knowledge acquisition} (learning new domain content); the
related capability is already present in the base.

The same pattern likely applies to other zero-base verifier benchmarks
in Table~\ref{tab:specialists} (Text2Cypher, LC-QuAD SPARQL): the base
produces well-formed prose or domain-adjacent structures that the strict
scorer rejects, and the specialist learns the target convention. We
flag this disclosure explicitly because the paper's headline gain on
LEDGAR is large ($+84.0$\,pp) and would otherwise be easily
misinterpreted as a much stronger capability claim than is warranted.

\section{Data, Dual-Use, and Compute Considerations}
\label{app:disclosures}

\paragraph{Data.}
All training data is from publicly available, license-permissive
sources. The PII adapter is trained on synthetic PII generated by
a templated tool; no real personal information is included in the
training set.

\paragraph{Dual-use.}
The same routing mechanism that selects between SQL and Cypher
experts could route between detection and generation for sensitive
categories. We release detection-oriented adapters only and
recommend deployment behind appropriate access controls.

\paragraph{Compute.}
Per-adapter training cost varies by track. Reasoning-trace adapters
on Qwen3.5-9B take roughly 3--6 GPU-hours each on a single L40S
(48\,GB); per-adapter wall times are reported in
Appendix~\ref{app:training}. Structured-generation specialists use
400--600 SFT steps; the chem/legal/quantum trio additionally goes
through an SFT-broad $\to$ SFT-narrow $\to$ GRPO curriculum, which
extends per-adapter wall time accordingly. All training and
evaluation runs were performed on a single L40S.

\end{document}